# An Intelligent Framework for Real-Time Yoga Pose Detection and Posture Correction

Chandramouli Haldar
Independent Researcher
NovaTech Innovative Solutions
Kolkata, India
0009-0004-9759-194X

*Abstract*— Yoga is widely recognized for its benefits in improving physical fitness, flexibility, and mental well-being; however, these benefits are highly dependent on the correctness of posture execution. Improper alignment during yoga practice can reduce effectiveness and increase the risk of musculoskeletal injuries, particularly in self-guided or online training environments. To address this challenge, this paper presents a hybrid Edge-AI–based framework for real-time yoga pose detection and posture correction. The proposed system integrates lightweight human pose estimation models with biomechanical feature extraction and a CNN–LSTM–based temporal learning architecture to accurately recognize yoga poses and analyze movement dynamics. Joint angles and skeletal features are computed from detected keypoints and compared with reference pose configurations to evaluate posture correctness. A quantitative scoring mechanism is introduced to assess alignment deviations and generate real-time corrective feedback through visual, textual, and voice-based guidance. Additionally, Edge-AI optimization techniques, including model quantization and pruning, are incorporated to enable low-latency performance on resource-constrained devices. The proposed framework aims to provide an intelligent, scalable, and accessible digital yoga assistant capable of enhancing user safety and training effectiveness in modern fitness applications.

## I. INTRODUCTION

Yoga is an ancient discipline that originated in India and has been practiced for centuries as a holistic method for improving physical health, mental well-being, and overall quality of life. It combines physical posture (asanas), breathing techniques (pranayama), and meditation to create harmony between the body and mind. Regular yoga practice has been associated with improved flexibility, muscle strength, cardiovascular health, and stress management. According to Rajendran and Sethuraman (2023), yoga also plays an important role in enhancing mental stability and supporting the management of various health conditions such as respiratory disorders, cardiovascular diseases, and depression. In the modern era, rapid urbanization, sedentary lifestyles, and increased work-related stress have significantly affected human health. Long working hours, excessive screen time, and reduced physical activity contribute to problems such as obesity, anxiety, and musculoskeletal disorders. As a result, yoga has gained global popularity as a simple and effective approach to maintaining physical fitness and mental balance. Its accessibility and ability to be practiced without complex equipment make it suitable for people of different age groups and fitness levels. However, the effectiveness of yoga largely depends on performing the postures correctly. Improper alignment during yoga poses can reduce the benefits of the exercise and may lead to injuries such as muscle strain, ligament damage, and joint stress. Vaishnnave and Manivannan (2025) emphasize that incorrect execution of yoga poses may cause musculoskeletal complications, particularly when practiced without proper supervision. Traditionally, yoga instructors guide practitioners in maintaining correct posture and breathing patterns, but the increasing use of online tutorials and self-guided learning has reduced direct supervision. This creates a need for systems that can monitor posture and provide real-time guidance.

Recent advancements in artificial intelligence (AI) and computer vision have created new opportunities for developing automated systems capable of analysing human body movements. Human Pose Estimation techniques enable computers to detect body joints and skeletal structures from images or videos, allowing systems to recognize and evaluate yoga poses. These technologies are widely used in modern fitness applications to monitor posture and provide feedback. Despite these technological advancements, performing yoga without a proper structural or analytical approach remains risky. Without accurate posture evaluation and corrective feedback, practitioners may unknowingly adopt incorrect body positions that could lead to injury. Therefore, integrating AI-based posture recognition systems with structured yoga training methods is essential to ensure safe and effective practice in modern digital fitness environments.

## II. LITERATURE REVIEW

Recent research in yoga posture recognition has increasingly focused on the application of artificial intelligence, computer vision, and machine learning techniques to assist practitioners in maintaining correct posture during yoga practice. Scholars widely agree that correct posture is fundamental to achieving the physical and therapeutic benefits of yoga while preventing injuries. Rajendran and Sethuraman (2023) emphasize that yoga contributes significantly to improving flexibility, muscular strength, cardiovascular health, and mental stability. However, they also note that improper posture during yoga practice can result in musculoskeletal strain and reduced effectiveness of the exercise. Similarly, Vaishnnave and Manivannan (2025) highlight that incorrect alignment during yoga poses may cause joint stress, ligament damage, and muscle injuries, thereby emphasizing the need for reliable posture monitoring systems. To address this challenge, researchers have explored several computational approaches for detecting and analyzing yoga poses. A commonly adopted technique is human pose estimation (HPE), which enables computers to detect body joints and represent the human body as a skeletal structure using keypoints extracted from images or videos. Rajendran and Sethuraman (2023) provide a comprehensive survey of yoga posture recognition systems and discuss various approaches including computer vision methods, machine learning models, deep learning architectures, and sensor-based monitoring systems. Their work establishes a foundational understanding of the technologies used in posture recognition and highlights the potential of AI-driven systems for remote fitness monitoring. However, as a survey study, it primarily focuses on reviewing existing approaches rather than proposing a new system or experimental framework.

Several studies have focused on evaluating different machine learning and deep learning techniques for yoga pose classification. Nayak et al. (2025) conducted a comparative analysis of multiple algorithms including Convolutional Neural Networks (CNN), Dense Neural Networks (DNN), Logistic Regression, Random Forest, Stochastic Gradient Descent, and Long Short-Term Memory (LSTM) networks. Using keypoints extracted through the BlazePose framework, the authors trained and tested these models on annotated yoga datasets. Their results demonstrated that Dense Neural Networks achieved the highest classification accuracy among the evaluated models. The study provides valuable insights into the comparative performance of various algorithms. However, the system primarily focuses on pose classification and does not extensively address posture correction or real-time feedback mechanisms.

Similarly, Tayal et al. (2025) investigated deep learning-based approaches for yoga pose detection by evaluating several convolutional neural network architectures including MobileNet, VGG19, and EfficientNet. Their research aimed to identify the most effective deep learning architecture for yoga pose classification. The results indicated that EfficientNet combined with the AdaDelta optimizer achieved the highest accuracy and F1-score. This study demonstrates the effectiveness of optimized deep learning architectures for improving pose detection accuracy. Nevertheless, the research mainly concentrates on classification performance and does not incorporate real-time posture correction or interactive user feedback systems.

While the previous studies emphasize classification accuracy, other researchers have focused on developing real-time guidance systems that can assist yoga practitioners during exercise. Gurav et al. (2025) proposed a computer vision-based yoga posture detection system using the MoveNet pose estimation model. The system calculates joint angles using cosine similarity and provides voice-based feedback to guide users in correcting their posture. Additionally, the system integrates a recommendation module that suggests yoga poses based on user flexibility and skill level. This approach demonstrates the potential of combining pose estimation with interactive feedback to create intelligent yoga assistant systems. However, the authors note that further improvements are required to enhance system accuracy and expand the dataset to include more yoga poses.

Pasalkar et al. (2025) introduced the PosePerfect system, which utilizes the MobileNetV2 deep learning architecture for real-time yoga pose detection and classification. The system is designed to function as an AI-powered yoga instructor by providing guided sessions, posture evaluation, and progress tracking. One advantage of this approach is its ability to offer interactive feedback and user engagement during practice. Nevertheless, the system may face limitations related to environmental variations such as lighting conditions and background complexity, which can affect pose detection accuracy in real-world scenarios.

Another significant contribution is presented by Zaghloul et al. (2025), who focused on developing a lightweight and smartphone-friendly pose recognition system. Their approach utilizes MediaPipe for body landmark detection combined with machine learning classification algorithms. By relying on landmark-based features rather than full image processing, the proposed system reduces computational complexity and enables real-time performance on mobile devices. This work highlights the importance of efficiency and accessibility in deploying AI-based fitness systems. However, the simplified approach may limit detection accuracy when compared to more complex deep learning models.

Overall, the reviewed studies demonstrate significant progress in the development of AI-based yoga posture recognition systems. While deep learning models such as CNN and EfficientNet provide high accuracy for pose classification, lightweight approaches enable real-time deployment on mobile devices. Additionally, recent research has begun to incorporate interactive feedback mechanisms to assist users during yoga practice. Despite these advancements, several challenges remain, including limited datasets, environmental variability, and the lack of comprehensive posture correction systems. Addressing these issues will be crucial for developing more robust and intelligent AI-assisted yoga training platforms in the future.

**Table 1. Comparative Table of Reviewed Studies**

| Study | Method/Technology | Key Contribution | Advantages | Limitations |
|---|---|---|---|---|
| Rajendran & Sethuraman (2023) | Survey of ML, DL, CV methods | Comprehensive review of yoga pose recognition techniques | Provides broad overview of methods and technologies | No experimental implementation |
| Vaishnnave & Manivannan (2025) | Machine learning + computer vision | Framework for yoga pose detection using ML algorithms | Explains key techniques for pose recognition | Limited experimental validation |
| Nayak et al. (2025) | CNN, DNN, RF, LR, LSTM with BlazePose | Comparative evaluation of ML and DL models | Systematic performance comparison | Focus only on classification accuracy |
| Tayal et al. (2025) | CNN architectures (MobileNet, VGG19, EfficientNet) | Performance analysis of deep learning models | High classification accuracy | No real-time feedback system |
| Gurav et al. (2025) | MoveNet + cosine similarity | Real-time posture detection with voice feedback | Interactive yoga guidance system | Limited dataset and robustness issues |
| Pasalkar et al. (2025) | MobileNetV2 | AI-based yoga instructor system | Provides guided sessions and progress tracking | Sensitive to environmental conditions |
| Zaghloul et al. (2025) | MediaPipe + ML classification | Smartphone-friendly pose recognition | Low computational cost, real-time performance | Slightly lower accuracy than deep learning models |

*Table 2. Algorithm Comparison Chart for Yoga Pose Detection Methods*

| Algorithm / Model | Used By | Purpose in System | Strengths | Weaknesses | Suitable Applications |
|---|---|---|---|---|---|
| **Convolutional Neural Network (CNN)** | Nayak et al. (2025); Tayal et al. (2025) | Image-based yoga pose classification | High accuracy in image recognition, strong feature extraction | Requires large datasets and computational power | Pose classification from images or videos |
| **Dense Neural Network (DNN)** | Nayak et al. (2025) | Classification using extracted skeletal features | Good performance with structured input data | Not ideal for spatial image features | Landmark-based pose classification |
| **Long Short-Term Memory (LSTM)** | Nayak et al. (2025) | Temporal pose sequence analysis | Captures motion patterns and time dependencies | Higher complexity and training time | Yoga movement or pose transition detection |
| **Random Forest (RF)** | Nayak et al. (2025) | Traditional machine learning classification | Easy to train, interpretable | Lower accuracy for complex pose recognition | Small dataset pose classification |
| **Logistic Regression (LR)** | Nayak et al. (2025) | Baseline classification model | Simple implementation, low computational cost | Limited capability for complex pose detection | Basic pose classification tasks |
| **MobileNet / MobileNetV2** | Tayal et al. (2025); Pasalkar et al. (2025) | Lightweight deep learning model for pose detection | Efficient for real-time applications and mobile devices | Slightly lower accuracy than larger networks | Mobile fitness applications |
| **VGG19** | Tayal et al. (2025) | Deep learning image classification | Strong feature extraction capability | Very computationally expensive | High-performance computing environments |
| **EfficientNet** | Tayal et al. (2025) | Optimized deep learning architecture | High accuracy with optimized parameters | Requires GPU resources | Advanced pose detection systems |
| **MoveNet Pose Estimation** | Gurav et al. (2025) | Real-time human pose estimation | Fast and efficient, supports real-time detection | Limited fine-grained posture correction | Real-time yoga monitoring systems |
| **MediaPipe Pose** | Zaghloul et al. (2025) | Landmark-based pose detection | Lightweight, mobile-friendly | Accuracy lower than deep CNN models | Smartphone-based yoga trainers |
| **BlazePose** | Nayak et al. (2025) | Keypoint extraction from human body | High speed and accuracy for skeletal detection | Requires further classification model | Pose detection preprocessing stage |

The comparative and algorithmic analysis summarized in Table 1 and Table 2 highlights the key methodological trends in existing AI-based yoga pose detection studies. Most of the reviewed works rely on human pose estimation combined with machine learning or deep learning models to recognize yoga postures from images or video data. As shown in Table 1, several studies focus primarily on improving classification performance using different learning algorithms. For instance, Nayak et al. (2025) conducted a comparative evaluation of multiple models—including CNN, Dense Neural Networks (DNN), Logistic Regression, Random Forest, Stochastic Gradient Descent, and LSTM—using skeletal keypoints extracted through the BlazePose framework. Their study reported that DNN achieved the highest accuracy among the tested models. Similarly, Tayal et al. (2025) investigated the effectiveness of different convolutional neural network architectures for yoga pose detection, including MobileNet, VGG19, and EfficientNet. Their experiments showed that EfficientNet produced the best performance when combined with the AdaDelta optimizer. These results reinforce the observation that deep learning architectures are currently dominant in image-based yoga pose classification tasks. The algorithm comparison summarized in Table 2 also shows that several studies employ pose estimation frameworks to extract body landmarks before classification. For example, Nayak et al. (2025) used BlazePose for keypoint extraction, Gurav et al. (2025) used the MoveNet model for real-time pose detection and joint-angle evaluation, and Zaghloul et al. (2025) implemented MediaPipe for landmark detection in a smartphone-based pose recognition system. These frameworks act as pre-processing components that convert human body movements into skeletal key points that can be analyzed by classification algorithms. Despite the progress demonstrated by these studies, several limitations are consistently observed. Many works rely on relatively limited yoga pose datasets, which restricts the ability of trained models to generalize to diverse real-world scenarios. Additionally, a large portion of the research focuses primarily on pose classification, while detailed posture evaluation or corrective feedback mechanisms are less extensively explored. Environmental factors such as lighting conditions, camera angles, and background variations can also influence the performance of computer vision-based pose detection systems. Furthermore, the analysis indicates a common challenge in balancing model accuracy with computational efficiency, particularly for systems intended to operate on mobile devices or in real-time environments.

### III. PROPOSED METHODOLOGY

To overcome the limitations observed in existing yoga posture recognition systems—such as limited dataset diversity, computational complexity, lack of temporal modelling, and insufficient real-time feedback—a Hybrid Edge-AI Based Yoga Pose Detection and Correction Framework is proposed. Prior studies have explored various machine learning and deep learning approaches for yoga posture classification; however, these systems often suffer from limited pose generalization, dependency on specific datasets, or high computational requirements (Nayak et al.; Tayal et al.). Furthermore, many systems rely solely on static pose classification without analysing temporal motion dynamics or biomechanical alignment, reducing their effectiveness for real-time posture correction (Gurav et al.; Pasalkar et al.). The proposed system integrates lightweight pose estimation models, biomechanical posture analysis, temporal deep learning architectures, and adaptive feedback mechanisms to improve both accuracy and efficiency. The framework consists of five main stages: data acquisition, pose estimation and key point extraction, biomechanical feature computation, temporal pose classification, and real-time posture evaluation and feedback generation.

*A. Data Acquisition and Pre-Processing*

The system captures input using a monocular RGB camera, such as a webcam or smartphone camera, which continuously records video frames while the user performs yoga poses. Each frame captured at time $t$ is represented as:

$$I_t \in R^{H \times W \times 3}$$

where $H$ and $W$ represent the spatial resolution of the frame and the three channels correspond to RGB color components.

To improve reliability of pose detection, pre-processing operations such as normalization, resizing, noise reduction, and background filtering are applied. The pre-processing function can be expressed as:

$$I'_t = f_{pre}(I_t)$$

where $I_t$ denotes the original frame and $I'_t$ represents the processed frame used for further analysis.

*B. Human Pose Estimation and Keypoint Detection*

After pre-processing, the system performs human pose estimation to detect body joints. Modern pose estimation frameworks such as MoveNet, BlazePose, and PoseNet allow real-time detection of body key points (Rajendran and Sethuraman; Gurav et al.). These models identify key anatomical landmarks including shoulders, elbows, wrists, hips, knees, and ankles.

The detected pose at time $t$ can be represented as a set of key points:

$$P_t = \{(x_i, y_i)\}_{i=1}^{N}$$

where $N$ represents the number of detected body key points and $x_i, y_i$ denotes the two-dimensional coordinates of the $i^{th}$ joint. These keypoints collectively form a skeletal representation of the human body, which is used as the basis for posture analysis. Lightweight architectures are preferred in order to maintain low inference latency and support deployment on edge devices (Zaghloul et al.).

### C. Biomechanical Feature Extraction

Although raw key point coordinates provide spatial information about body posture, they are insufficient to accurately evaluate posture correctness. Therefore, the system extracts biomechanical features, including joint angles, limb orientations, and body alignment parameters.

Consider three key points $A, B,$ and $C$ representing adjacent body joints. The vectors between these joints are defined as:

$$\overrightarrow{BA} = A - B$$
$$\overrightarrow{BC} = C - B$$

The joint angle at point $B$ is computed using the cosine similarity formula:

$$\theta = \cos^{-1}\left(\frac{\overrightarrow{BA} \cdot \overrightarrow{BC}}{|\overrightarrow{BA}| \, |\overrightarrow{BC}|}\right)$$

where $\theta$ represents the angle between the two body segments. Using this approach, multiple joint angles are computed to form a biomechanical feature vector:

$$F_t = [\theta_1, \theta_2, \theta_3, \ldots, \theta_k]$$

where $k$ denotes the number of extracted joint angles.

Biomechanical feature extraction improves posture recognition by reducing the effect of camera distance, body proportions, and environmental noise, which are known limitations in vision-based systems (Vaishnnave et al.).

### D. Temporal Pose Modeling

Yoga poses often involve dynamic movements and transitions, which cannot be effectively analyzed using single-frame approaches. To capture temporal dynamics, the proposed system incorporates temporal deep learning models.

A CNN–LSTM architecture is used to analyze pose sequences. Given a sequence of feature vectors:

$$S = \{F_1, F_2, \ldots, F_T\}$$

where $T$ represents the number of frames in the sequence, the temporal relationship between poses is modeled using an LSTM network:

$$h_t = \text{LSTM}(F_t, h_{t-1})$$

where $h_t$ denotes the hidden state of the network at time $t$.

The final pose classification output is obtained using the SoftMax function:

$$P(y|S) = \text{Softmax}(Wh_T + b)$$

where $W$ and $b$ represent the weight matrix and bias vector, respectively, and $y$ denotes the predicted yoga pose class.

Temporal modelling allows the system to capture motion patterns associated with specific yoga poses and improves classification accuracy (Tayal et al.).

### E. Pose Similarity and Posture Evaluation

Once the yoga pose is classified, the system evaluates posture correctness by comparing the detected joint angles with reference angles stored in a pose database.

Let:

$$\theta_i$$

represent the detected joint angle and
$$\theta_i^{ref}$$
represent the corresponding reference joint angle. The angular deviation is calculated as:
$$\Delta_i = |\theta_i - \theta_i^{ref}|$$
The overall posture accuracy score is then computed as:
$$Score = \frac{1}{N}\sum_{i=1}^{N}\left(1 - \frac{\Delta_i}{\theta_{max}}\right)$$
where $N$ is the number of evaluated joints and $\theta_{max}$ represents the maximum allowable deviation. If the deviation exceeds a predefined threshold, the corresponding joint is marked as incorrect and corrective guidance is generated. This scoring mechanism enables quantitative evaluation of posture accuracy and addresses limitations of simple threshold-based systems reported in previous studies (Nayak et al.).

### F. Real-Time Feedback Generation

Based on the evaluation results, the system provides real-time corrective feedback to guide the user toward proper posture alignment. Three feedback mechanisms are implemented:
1. Visual feedback, where a skeletal overlay highlights incorrectly aligned joints.
2. Text-based guidance, which provides instructions such as adjusting arm or leg position.
3. Voice feedback, allowing hands-free posture correction during yoga practice.

Real-time feedback significantly improves usability and learning efficiency in AI-based yoga training systems (Pasalkar et al.; Gurav et al.).

### G. Edge-AI Optimization

Deep learning models typically require substantial computational resources, limiting their deployment on mobile devices. To address this challenge, the proposed framework incorporates Edge-AI optimization techniques, including model quantization, network pruning, and knowledge distillation. These optimization strategies reduce model size and inference latency while maintaining accuracy, enabling real-time operation on mobile and edge computing platforms (Zaghloul et al.).

### H. System Workflow

The complete workflow of the proposed system can be summarized as follows:
1. Capture video frames from the camera.
2. Apply pre-processing to improve image quality.
3. Detect body key points using a lightweight pose estimation model.
4. Extract biomechanical features such as joint angles.
5. Model temporal pose dynamics using a CNN–LSTM architecture.
6. Classify yoga poses.
7. Evaluate posture correctness using similarity metrics.
8. Generate real-time corrective feedback.

This integrated framework combines accurate pose detection, temporal modelling, biomechanical evaluation, and efficient edge deployment, addressing key limitations identified in existing yoga posture recognition systems.

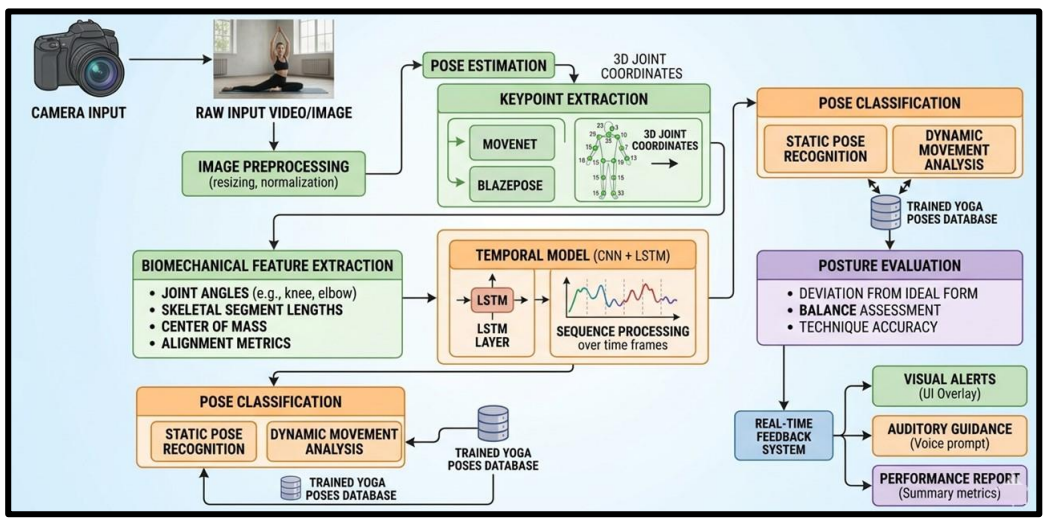

*Figure 1. Proposed Yoga Pose Detection System Architecture*

## IV. EXPERIMENTAL SETUP

To evaluate the performance of the proposed Hybrid Edge-AI Yoga Pose Detection and Correction System, experiments were conducted using a combination of publicly available yoga pose datasets and real-time webcam input. Public datasets such as Yoga-82 and Yoga Pose Image Dataset were used to train and test the pose classification model, while additional video samples were captured using a standard RGB camera to evaluate real-time performance. The dataset was divided into training (70%), validation (15%), and testing (15%) sets to ensure reliable evaluation. The system was implemented using Python, with libraries including TensorFlow/PyTorch, OpenCV, and MediaPipe or MoveNet for pose estimation. Experiments were performed on Google Colab for model training and Testing. The proposed architecture employs a CNN–LSTM model to capture both spatial and temporal pose features, trained using the Adam optimizer with a learning rate of 0.001, batch size of 32, and 50 training epochs. System performance was evaluated using standard metrics including accuracy, precision, recall, and F1-score, while real-time capability was measured through frame processing rate and inference latency. Additionally, posture correctness was evaluated by comparing detected joint angles with reference pose angles, where the deviation between angles was computed and used to generate a posture accuracy score, enabling the system to provide real-time corrective feedback during yoga practice.

## V. IMPLEMENTATION FRAMEWORK

The proposed yoga pose detection system is designed as a modular framework that can be implemented using widely available computer vision and deep learning tools. The system relies on RGB video input captured through a webcam or smartphone camera. The captured frames are processed using pose estimation models such as MoveNet, BlazePose, or MediaPipe to extract human body keypoints. The extracted skeletal keypoints are then used to compute biomechanical features including joint angles and limb orientations. These features form the input to a temporal deep learning model based on a CNN–LSTM architecture, which is capable of analyzing pose sequences and recognizing yoga postures. Once a pose is classified, the system evaluates posture correctness by comparing detected joint angles with reference pose angles stored in a database. The framework is designed to support real-time operation on edge devices using model optimization techniques such as quantization and pruning. Although this study focuses on the design of the system architecture, future implementations can evaluate the framework using publicly available yoga pose datasets such as Yoga-82 or Yoga Pose Image Dataset. Performance evaluation could be conducted using metrics including classification accuracy, precision, recall, and inference latency.

## VI. FUTURE WORK

Future work will focus on implementing the proposed framework and evaluating its performance using publicly available yoga pose datasets. Experimental studies will analyze the accuracy of the pose classification model, the reliability of the posture evaluation module, and the real-time performance of the system on edge devices. Additionally, further improvements may include integrating 3D pose estimation techniques, expanding the yoga pose dataset, and incorporating personalized training recommendations for users.

## VII. CONSLUSION

This paper presented a proposed AI-based yoga posture detection and correction framework aimed at assisting practitioners in performing yoga poses accurately and safely. The study reviewed existing research on yoga pose recognition and identified several limitations in current systems, including limited datasets, lack of temporal analysis, insufficient posture correction mechanisms, and computational constraints for real-time applications. To address these challenges, a Hybrid Edge-AI Yoga Pose Detection System was proposed that integrates human pose estimation, biomechanical feature extraction, and temporal deep learning models within a unified architecture. The proposed framework utilizes pose estimation models such as MoveNet, BlazePose, or MediaPipe to detect body key points, which are then used to compute joint angles and biomechanical features for posture analysis. A CNN–LSTM based temporal model is incorporated to analyze pose sequences and improve classification reliability. The system further evaluates posture correctness by comparing detected joint angles with reference poses and generates real-time visual, textual, and voice-based feedback to guide users in correcting their posture. Additionally, Edge-AI optimization techniques are considered to enable efficient deployment on mobile and real-time environments. Although the current work focuses on the conceptual design of the system architecture, the proposed framework provides a foundation for developing intelligent AI-assisted yoga training platforms. Future work will involve implementing the system, evaluating its performance using yoga pose datasets, and exploring advanced techniques such as 3D pose estimation and personalized yoga recommendations to further enhance system capabilities.